\documentclass{article}\usepackage[hidelinks]{hyperref}

\usepackage{arxiv}

\usepackage[utf8]{inputenc} 
\usepackage[T1]{fontenc}    
\usepackage{hyperref}       
\usepackage{url}            
\usepackage{booktabs}       
\usepackage{amsfonts}       
\usepackage{nicefrac}       
\usepackage{microtype}      
\usepackage{lipsum}		
\usepackage{graphicx}
\usepackage[square,numbers]{natbib}
\usepackage{doi}

\usepackage{graphicx}
\usepackage[utf8]{inputenc}
\usepackage[english]{babel}
\usepackage{cancel}
\usepackage{amsmath}
\usepackage{url}
\usepackage{multirow}
\usepackage{array,multirow}
\usepackage{algorithm}
\usepackage{algorithmic}
\usepackage{graphicx}
\usepackage{verbatim}
\graphicspath{ {./figure/} }
\usepackage{babel}
\usepackage{xfrac}
\usepackage{dblfloatfix}
\usepackage{etoolbox}
\makeatletter
\patchcmd{\@makecaption}
  {\scshape}
  {}
  {}
  {}
\makeatother
\usepackage{xcolor}
\usepackage{tabularx}
\usepackage{amssymb}
\usepackage{amsthm}
\usepackage{array, boldline, makecell, booktabs}

\newtheorem{definition}{Definition}

\newtheorem{Corollary}{Corollary}
\newtheorem{theorem}{Theorem}
\usepackage{multirow}
\usepackage{enumitem}
\usepackage{diagbox}
\usepackage{afterpage}

\usepackage{color,soul}
\usepackage{array,multirow}
\usepackage{caption}

\usepackage{microtype}
\usepackage{booktabs}
\usepackage[format=hang,font={small,bf}]{caption}
\usepackage[utf8]{inputenc}
\usepackage{color,soul}
\setul{0.5ex}{0.3ex}
\definecolor{Green}{rgb}{0,1,0}
\setulcolor{Green}
\usepackage[bottom]{footmisc}

\usepackage{nccmath}
\usepackage{mathtools}

\title{Defending Active Directory by Combining Neural Network based Dynamic Program and Evolutionary Diversity Optimisation}

\author{Diksha Goel\\ School of Computer Science \\ University of Adelaide, Australia \\diksha.goel@adelaide.edu.au
\And
Max Ward\\ Department of Molecular and Cellular\\Biology, Harvard University, USA \\ School of Computer Science\\ University of Adelaide, Australia \\maxwardgraham@fas.harvard.edu
\And
Aneta Neumann\\ School of Computer Science\\University of Adelaide, Australia \\aneta.neumann@adelaide.edu.au
\And
Frank Neumann\\ School of Computer Science \\University of Adelaide, Australia\\frank.neumann@adelaide.edu.au
\And
Hung Nguyen\\School of Computer Science\\University of Adelaide, Australia\\hung.nguyen@adelaide.edu.au
\And
Mingyu Guo\\ School of Computer Science\\University of Adelaide, Australia\\mingyu.guo@adelaide.edu.au}

\begin{document}
\maketitle
\begin{abstract}
Active Directory (AD) is the default security management system for Windows domain networks. We study a Stackelberg game model between one attacker and one defender on an AD attack graph. The attacker initially has access to a set of entry nodes. The attacker can expand this set by strategically exploring edges. Every edge has a detection rate and a failure rate. The attacker aims to maximize their chance of successfully reaching the destination before getting detected. The defender's task is to block a constant number of edges to decrease the attacker's chance of success. We show that the problem is \#P-hard and, therefore, intractable to solve exactly. We convert the attacker's problem to an exponential sized Dynamic Program that is approximated by a Neural Network (NN). Once trained, the NN provides an efficient fitness function for the defender's Evolutionary Diversity Optimisation (EDO). The diversity emphasis on the defender's solution provides a diverse set of training samples, which improves the training accuracy of our NN for modelling the attacker. We go back and forth between NN training and EDO. Experimental results show that for R500 graph, our proposed EDO based defense is less than 1\% away from the optimal defense.
\end{abstract}




\section{Introduction}
Cyber attack techniques utilize attack graphs to identify the possible ways an attacker can exploit to gain unauthorized access to the systems. 
Industrial practitioners are actively using the Active Directory attack graph, which is an attack graph model. Microsoft \textit{Active Directory} is a security control system for Windows domain networks \cite{dias2002guide}. 
Microsoft domain network constitutes significant market shares among small and big organizations globally, due to which Active Directories are promising targets for cyber attacks. 
AD structure depicts an attack graph where nodes are the computers, accounts, applications, and an edge from node X to node Y denotes that an attacker may access node Y from node X using some existing access or known exploits. Various applications and tools are designed to investigate the AD attack graphs; however, BLOODHOUND\footnote{https://github.com/BloodHoundAD/BloodHound} is the most popular tool. 

\begin{figure}[t!]
\centering
 \includegraphics[width=0.6\paperwidth]{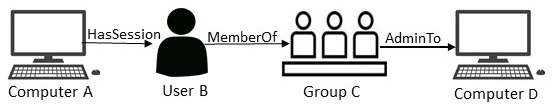}
 \caption{Example of BLOODHOUND snowball identity attack.}
 \label{fig:bloodhound}
\end{figure}

BLOODHOUND models an \textit{identity snowball attack} where the attacker initiates an attack by gaining access to a low privileged account (attacker’s entry node) via a phishing attack and tries to move to other nodes aiming to reach the most-privileged account, DOMAIN ADMIN (DA) \cite{dunagan2009heat}. Figure \ref{fig:bloodhound} illustrates an example of BLOODHOUND snowball identity attack. One fundamental functionality of BLOODHOUND is that it uses Dijkstra’s algorithm to generate the shortest attack paths from the entry node of the attacker to the DA and measures the length of attack path in number of hops. BLOODHOUND has made it much easier for the attacker to attack AD. Due to the popularity of AD attack graphs, defenders also study AD graphs for designing defensive strategies. Dunagan et al. \cite{dunagan2009heat} proposed a heuristic solution for blocking a few edges to partition the attack graph into various disconnected components. The resulting disconnected components of the graph disable attackers from reaching DA only if the attacker’s entry nodes and DA are in different components. 

Evolutionary algorithms have traditionally been used to solve various attacker-defender problems \cite{zhang2019investigating, hu2020optimal, hemberg2018adversarial}. Ulrich et al. \cite{ulrich2010integrating, ulrich2011maximizing} studied Evolutionary Diversity Optimisation (EDO) that aims to find a set of diverse solutions. EDO has gained considerable attention in the Evolutionary Computation community. A new solution deviating from its predecessors leads to less competitiveness and high evolvability \cite{lehman2013evolvability}. \textit{\textbf{In this paper, the defender aims to block a set of edges to minimise the strategic attacker’s probability of reaching DA.}} To solve the defender’s problem, we consider blocking plans of the defender as a population and propose EDO based defensive approach to generate a diverse set of blocking plans.

In the literature, Guo et al. \cite{guo2021practical} proposed several algorithms to address this problem. However, the authors have considered a scenario where once the attacker is detected, the attack ends and when an attacker chooses a path, the attacker continues to move on to that path until the attacker gets caught or reaches DA. The attacker's problem in \cite{guo2021practical} can be solved using Dijkstra's algorithm, but in our model, the attacker's problem is computationally hard. In our model, we assume that every edge $e$ has a different \textit{detection probability $p_{d(e)}$}, i.e., if the attacker is detected, the whole attack ends and a \textit{failure probability $p_{f(e)}$}, i.e., if the attacker fails and is not able to pass through an edge; the attacker can try another edge from the same node or different node. For e.g., an edge requires cracking a password to pass through it; if the password is weak, then the attacker can crack it. In this case, the probability of having a strong password is the failure probability. With every edge’s $p_{d(e)}$ and $p_{f(e)}$, the attacker can successfully go through an edge $e$ with a \textit{success probability} $p_{s(e)}=(1-p_{d(e)} - p_{f(e)})$. 

\textit{Therefore, in our model, the attacker plays strategically in the sense that the attacker first starts an attack, and if, at some point, fails (instead of being detected), the attacker can try again until the attacker is detected or has tried all possible options. The attacker aims to design an attacking policy that maximises their probability of reaching DA.} Initially, the attacker has access only to a set of entry nodes, from where the attacker can start. While trying to reach DA, the attacker expands the set of accessible nodes by exploring edges and saves this information, i.e., set of successful edges the attacker has currently control of, set of edges that the attacker has lost (failed but not detected, for not being able to guess the password, etc.). All the strategies that attacker has tried are an “investment” that the attacker can use later. In this way, the attacker has \textit{“secured”} a set of nodes at any point and can attempt an unattempted edge originating from any of the secured nodes or entry nodes, which we combinedly call checkpoints, \textit{$\text{Checkpoints} = \{\text{Entry nodes} \cup \text{Secured nodes}\}$}. Generally, the attacker prefers the attack paths with low detection and failure probability, and covers the valuable checkpoints along its way. 
The defender wants to deterministically block $k$ \textit{block-worthy edges}, where $k$ is the defender’s budget, in turn increasing the corresponding edge’s failure rate from original $p_f$ to 100\%. We follow a standard \textit{Stackelberg game model} \cite{yin2010stackelberg}, where 
the defender devises a strategy, and the attacker observes the defender’s strategy and plays his best to develop an attack strategy on the target. In practice, the attacker can scan the AD environment using SHARPHOUND\footnote{https://github.com/BloodHoundAD/SharpHound} tool and get information about which edges are blocked. 

\textit{We aim to propose an effective approach for computing defender strategy (to block a set of edges that minimises strategic attackers’ probability of reaching DA) that scales to large AD attack graphs.} We have proved that the attacker’s problem of deriving an optimal attacking policy is $\#P$-hard. We proved that the defender’s problem is also $\#P$-hard, even if the blocking budget is one. Therefore, the problem is intractable to solve exactly with the current methods, so we propose an approach that involves training a \textbf{\textit{Neural Network (NN)}} to approximate attacker's problem and \textbf{\textit{Evolutionary Diversity Optimisation}} to solve the defender's problem. We can describe the attacker’s problem as Markov Decision Process (MDP), which can be solved using Dynamic Program (DP) \cite{bellman1966dynamic}. The size of state space is $3^{\text{|Edges|}}$, where an edge is either unattempted, attempted and failed, or attempted and successful. However, this state space is too large considering that practical AD graphs may have tens of thousands of edges. We use a \textit{fixed-parameter analysis technique} that focuses on determining easy-to-solve instances of a problem. 

We propose kernelization technique that exploits the structural features of the AD attack graphs to obtain a much smaller \textit{condensed graph} and convert the problem from a condensed graph to a DP. Using our kernelization technique, the state space becomes Fixed Parameter Tractable with respect to a parameter called the number of \textit{Non-Splitting Paths (NSP)}. Guo et al. \cite{guo2021practical, guo2022scalable} proposed NSP idea to describe tree-likeness of AD graphs. NSP is a path on which every node has only one successor except the last node. For small AD graphs, we can now solve DP and for larger AD graphs, our FPT special parameter $\#NSP$ is practically too large, so we use NN to approximate the DP\cite{yang2018boosting}. Our main idea is to train NN to learn both the base cases and the DP recursive relationships. Considering that the state space is exponential and we do not have resources to train NN to learn the value of every state; however, not all states are useful. It is important to learn the values of the states that are referenced by the optimal decision path, therefore, we only consider the important states in the state space, which reduces the state space size to a large extent. With the strong flexibility power of NN, we aim to train the NN to approximate the attacker's policy. Once trained, NN acts as an efficient fitness function for EDO (the exact fitness function is $\#$P-hard to compute). EDO provides a diverse set of blocking plans, i.e., diverse set of training samples for training NN, which prevents NN from getting stuck in local optimum. The blocking plans are given as input to the NN, and it outputs the attacker's probability of reaching DA corresponding to the blocking plan. We go back and forth between the processes, generating blocking plans using EDO and training NN on blocking plans, to get a well trained NN that can act as an efficient fitness function for EDO. \textit{In this way, EDO and NN help each other in order to improve.}
 
\noindent \textbf{Our contributions.} We first prove that the attacker’s problem of deriving an optimal attacking policy is $\#P$-hard, and the defender’s problem is also  $\#P$-hard. We then design a kernelization technique that converts the attacker’s problem to a dynamic program where the number of states is fixed-parameter tractable with respect to the number of non-splitting paths. For small AD graphs, we can directly solve the attacker's problem. For larger graphs, we train a Neural Network to approximate the Dynamic Program. We propose Evolutionary Diversity Optimisation to solve the defender’s edge blocking problem. Our experimental results on the synthetic R500 AD attack graph show that the proposed approach (attacker’s policy and defender’s policy) is highly effective, and is less than 1\%  away from the optimal solution.

\section{RELATED WORK}
\noindent \textbf{\textit{AD Attack Graphs.}} Guo et al. \cite{guo2021practical} investigated the edge interdiction problem for AD attack graphs where the defender's goal is to maximise the expected path length of the attacker. The authors proposed a Non-Splitting Path idea to describe the tree-likeness of AD attack graphs. The authors used Graph Neural Networks to determine attacker's local decision at each split node, where when an attacker chooses a path, the attacker continues to move on to that path until the attacker reaches a split node (node with multiple outgoing edges), and the attacker needs to decide which outgoing edge to choose. However, in our model, the attacker may have many checkpoints across the whole graph, and the attacker can explore the graph from any checkpoint. Therefore, the attacker makes a global decision by considering a set of split nodes. Due to this reason, the solution  \cite{guo2021practical} is not applicable to our model. 

\noindent \textbf{\textit{Neural Networks.}} 
Yang et al. \cite{yang2018boosting} proposed a general approach to boost the performance of Dynamic Program with Neural Networks. The authors used a solution reconstruction procedure that samples solutions, and a NN is trained to assess the quality of each solution. Xu et al. \cite{xu2020deep} designed a model that integrates Neural Networks with Dynamic Program to solve various optimisation problems. The authors proposed two solutions, value and policy-based, which considerably reduce time with reasonable performance loss. 

\noindent \textbf{\textit{Evolutionary Diversity Optimisation.}} 
Identifying high-quality, diverse solutions has attained huge attention in Evolutionary Computation under the concept of Evolutionary Diversity Optimisation \cite{ulrich2010integrating}, and Quality Diversity \cite{cully2017quality}. Both approaches have recently been applied to solve traditional combinatorial optimisation problems. Hebrard et al. \cite{hebrard2005finding} proposed a solution to find diverse solutions for constrained programming. Do et al. \cite{do2020evolving} studied the computation of high-quality and diverse solutions for the travelling salesman problem. In another study, Do et al. \cite{do2021analysis}  analysed EDO techniques for various permutation problems. In this paper, we utilise EDO to solve defender's problem by obtaining a diverse set of edge blocking plans for the $\#P$-hard attacker-defender Stackelberg game. In addition, we use trained NN as a fitness function for the defender’s EDO.

\section{Model Description}
AD attack graph is a directed graph $G=(V,E)$, with $n=|V|$ nodes and $m=|E|$ edges. There are $s$ entry nodes, from where the attacker can enter the graph and one destination node DA (Domain Admin). AD attack graphs may have multiple admin nodes, but we merge all admin nodes into one node and call it DA. The attacker starts from any entry node and aims to devise a policy that maximizes their probability of reaching DA. The attacker initially has access only to a set of entry nodes and tries to expand this set by exploring more edges. Every edge $e \in E$ has a detection probability $p_{d(e)}$ that ends the attack, and a failure probability $p_{f(e)}$, which does not end the attack; on encountering a failed edge, the attacker can continue the attack by exploring one of the unexplored edges. The attacker can successfully pass through an edge with a probability of $(1-p_{d(e)} - p_{f(e)})$. In our model, the attacker plays strategically; the attacker initiates an attack and continues the attack by exploring unexplored edges until the attacker is detected, has explored all possible options or reached DA. The defender blocks $k$ block-worthy edges, where $k$ is the defender's budget and aims to minimize the strategic attacker's chances of reaching DA. Only a set of edges are blockable. We assume that the attacker can observe the defensive action and accordingly comes up with the best-response attacking policy. We first show that the attacker’s and defender’s problems are $\#P$-hard to calculate. Therefore, the problems are intractable to solve with the existing approaches. We propose an approach that trains a Neural Network to approximate the attacker’s problem and Evolutionary Diversity Optimisation to solve the defender’s problem.
\begin{theorem}
\label{t_1}
\normalfont The attacker's value is $\#$P-hard to calculate.
\end{theorem}
\begin{theorem}
\label{t_2}
\normalfont The attacker's optimal policy is $\#$P-hard to calculate.
\end{theorem}
\begin{Corollary}
\label{t_3}
\normalfont The defender's value and policy are both $\#$P-hard to calculate.
\end{Corollary}
Due to space constraint, the detailed proofs of Theorems and Corollary are deferred to the full version of this paper. Our proofs are based on showing that the problem of calculating the attacker's value and policy, and defender's value and policy are at least as hard as the $\#P$-complete $s-t$ connectedness problem \cite{valiant1979complexity}.

\section{PROPOSED METHODOLOGY}
This section first discusses the proposed kernelization technique that reduces the original graph to a condensed graph and converts the attacker’s problem to a dynamic program. We then discuss the proposed Neural Network that approximates the attacker’s problem. Lastly, we discuss the proposed Evolutionary Diversity Optimisation to design the defensive policy.

\subsection{Fixed-Parameter Tractable Kernelization}
\noindent \textit{Kernelization} is a common technique for fixed-parameter analysis that processes a problem and reduces it to a smaller equivalent problem, called \textit{kernel}. 
We can consider an attack graph as a tree with $h$ extra edges, known as \textit{feedback edges (h)} and $h = m-(n-1)$. \textit{Splitting nodes (t)} are the nodes with more than one outgoing edge. DA does not have any successor in the graph, and even if DA have any successors, that can be ignored as once the attacker reaches DA, the attack ends. When $t = 0$, the attack graph is exactly a tree. Also,  $t\leq h$. We denote the \textit{Number of entry nodes} using $s$. For kernelization, we consider one parameter of AD attack graphs, which is the number of Non-Splitting Paths (paths from entry nodes, or other splitting nodes).
\begin{definition}
\label{def_NSP}
\noindent \textbf{Non-Splitting Path (NSP).}  Non-Splitting Path NSP$(i, j)$ is a path that goes from node $i$ to $j$, where $j$ is $i's$ successor and then continually moves to $j's$ sole successor, until we reach DA or another splitting node \cite{guo2021practical, guo2022scalable}.
\end{definition}
\begin{equation*}
NSP = \{NSP(i,j)|\, i \in \text{SPLIT} \cup \text{ENTRY}\} 
\end{equation*}
where SPLIT represents the set of splitting nodes and ENTRY represents the set of entry nodes. A NSP is \textit{blockable} only if at least one of its edge is blockable. In addition, once the attacker chooses to move onto a NSP, it is without the loss of generality to assume that under the attacker's optimal policy, the attacker has to complete the NSP until the attacker 1) gets caught; 2) fails; 3) reaches DA; or 4) reaches any splitting node. Otherwise, the attacker risks getting detected without securing any new checkpoints (splitting nodes).
\begin{figure}[h!]
\centering
 \includegraphics[width=0.4\paperwidth]{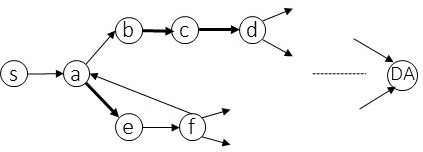}
 \caption{Example of attack graph.}
 \label{fig:attack_graph_nodes}
\end{figure}
\begin{definition}
\noindent \textbf{Block-worthy (BW).} A block-worthy $bw(i, j)$ is any furthermost blockable edge on path NSP$(i,j)$. Two NSPs may share the same block-worthy edge. 
\end{definition}
\begin{equation*}
BW = \{bw(i,j)|\, i \in \text{SPLIT} \cup \text{ENTRY}, j \in \text{Successor$(i)$}\} 
\end{equation*}

Notably, for the edge blocking strategy, we only need to spend one unit of budget on $NSP(i,j)$; otherwise, we could simply block $bw(i,j)$ to eliminate this NSP from the attacker's consideration. 
Size of block-worthy edge set, $|BW|$ can be bounded as:
\begin{align*}
\begin{split}
    |BW|& \leq s+t+h\\
    |BW|& \leq s + 2h,\,\,\,\,\,\,\,\,\, \text{since}\,\, t \leq h
\end{split}
\end{align*}

Figure \ref{fig:attack_graph_nodes} illustrates an attack graph, where the entry node is $s$, destination node is $DA$, split nodes are $\{a, d, f\}$,  non-splitting paths are $\{(s, a), (a, b, c, d), (a, e, f)\}$. The thick edges are blockable, so blockable edges are $\{(b, c), (c, d), (a, e)\}$ and block-worthy set is $\{(c, d), (a, e)\}$. \textit{In the \textbf{original AD attack graph}, there are $n$ nodes and $m$ edges. The kerneization technique converts the original graph into a \textbf{condensed AD attack graph} with only (|ENTRY|+|SPLIT| + 1) nodes and |NSP| edges.}\\

\noindent \textbf{Converting Attacker's Problem to Dynamic Program.} 
\noindent We describe attacker's problem of devising an attacking policy that maximises the probability of reaching DA as Markov Decision Process, where the state $s$ is a vector of size $|NSP|$ and can be represented as:

\begin{equation}
\label{attacker state vector}
\underbrace{< S, F, ?, ?, ?, S, F, ?, ?, ?, ?, ?, S, F, F >}_\text{Length of state vector = Number of NSP} 
\end{equation}
where:
\begin{description}
\item `S' represents that the attacker has tried NSP and it is successful (the attacker has reached at the end of NSP)
\item `?' represents that the attacker has not yet attempted the NSP
\item `F' represents that attacker has tried NSP and failed (not detected) 
\end{description}

Given a state vector as shown in Eq. (\ref{attacker state vector}), the attacker tries one of the NSP (action for a state) with status `?' (not attempted) in order to reach DA. The realisation of the NSP that the attacker tries can turn out to be either successful, fail or detected. Accordingly, the status of that NSP changes to `S' or `F' and the attacker gets a new state. However, if the attacker is detected, the attack ends. All the NSPs that the attacker has tried and are successful act as checkpoints for the attacker. The attacker can explore an unexplored edge from these checkpoints in the future. In this way, the attacker tries unexplored NSPs until the attacker reaches DA or is detected.\\

\begin{figure}[h!]
 \centering
 \includegraphics[width=0.3\paperwidth]{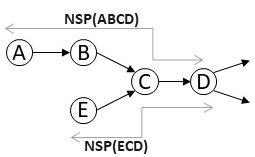}
 \caption{Example of state transition process.}
 \label{fig:state}
\end{figure}

\noindent \textbf{State Transition.} For a state and action (actions for a state s are the unexplored NSPs outgoing from the end node of successful NSPs in state s), we may have a distribution of future states. We have described the state transition process using Figure \ref{fig:state}. For simplicity, assume that every edge in Figure \ref{fig:state} has $p_d$ = 0.1 and $p_f$ = 0.2. We have two NSPs, $<$NSP(ABCD), NSP(ECD)$>$ visible in our figure, so we focus on these two NSPs for presentation purposes. The initial state vector is $\,\,<?, ?>$. We can try one of the NSPs. Let us try NSP(ABCD), and we go through each edge on this NSP. 
If AB fails: the status of its NSP changes to `F', and the new state is <F, ?>. This state is added to the future state set with 0.2 transition probability. There is 0.1 probability the attack ends and 0.7 probability of successfully passing through edge. If edge AB succeeds but BC fails: <F,?> state is already in set; therefore, its transition probability is updated to (0.7$\times$0.2+0.2)=0.34. There is (0.7$\times$0.1)=0.07 probability that the attack ends and the probability of successfully passing through edge is (0.7$\times$0.7)=0.49. If edge AB succeeds, BC succeeds, but CD fails: CD is a common block-worthy edge between NSP(ABCD) and NSP(ECD). If CD fails, both the NSPs sharing CD fail, and the new state is <F, F>. The state is added to the future state set with (0.7$\times$0.7$\times$0.2)= 0.098 transition probability. There is (0.7$\times$0.7$\times$0.1)=0.049 probability that the attack ends, and the probability of successfully passing through edge is (0.7$\times$0.7$\times$0.7)=0.343. If AB succeeds, BC succeeds, CD succeeds: All the edges on NSP(ABCD) have been explored, and we can successfully go through this NSP. Therefore, we have a new state <S, ?>. This state is added to the future state set with 0.343 transition probability. On trying NSP(ABCD), we get three future states and their corresponding transition probabilities, $\{(<F, ? >,\;0.34),(<F, F >,\;0.098), (<S, ? >,\;0.343)\}$.\\

During state transition, for each new state $s$, we first check if it is already a determined state or not. Determined states are the states which are already present in future state set and for each determined state, we have identified set of actions that can be performed on that state. If the state is not determined yet, we compute \textbf{\textit{admissible set of actions}} $A(s)$, for this state. Admissible set of actions are the actions available for the state, i.e., set of unexplored NSPs that can be explored from the checkpoints of this state. Notably, admissible actions only include unexplored NSPs and ignore those NSPs for which we do not have access to their source node. The maximum size of the attacker’s state space can be $3^{|NSP|}$, which is very large; however, not all state vectors are possible or relevant for the attacker. We only consider the state vectors that are relevant for the attacker by following the state transition process; given an initial state and admissible set of action, we determine the future states that an attacker can encounter on the way to DA, and we call these states as important states. 
We can solve the attacker’s problem using the DP technique. 
For a given state $s$ and action $a$ from \textit{admissible set of actions} $a \in A(s)$, the attacker problem of maximizing the probability of reaching DA can be written as:
\begin{equation}
\label{DP}
V(s) = \max_{\substack{a\in A(s)}} \,\bigg(\sum_{\substack{s'}}{Pr(s'\,|\,s,a)\, V(s')}\bigg)
\end{equation}
where $V(s)$ denotes the value function for the current state $s$ i.e., probability of reaching DA when the attacker is in state $s$, $s'$ denotes the distribution of future states that follows after choosing action $a$ and $Pr(s'\,|\,s,a)$ denotes the state transition  probability of $s'$, when an action $a$ is performed on $s$. Eq. (\ref{DP}) can be solved by computing the value functions for smaller problems and the overall value function step by step. 
Nevertheless, backward induction can be computationally challenging for large state spaces. Therefore, we use NNs to approximate the dynamic programming function.

\subsection{Attacker's Policy: Neural Network based Dynamic Program (NNDP)}
We use NN to approximate the attacker’s problem, and it acts as a fitness function for the Evolutionary Diversity Optimisation. Given a blocking plan, NN outputs a value that indicates the attacker's chances of reaching DA. For approximating the attacker's value function, we first supervise NN to learn the DP base states. Base states are the states in which the status of NSPs that end at DA is either `S' or `F'. NSPs that ends at DA with status `S' are always 100\% successful, and therefore, the value of these states is 1; the attacker will reach DA. If all the NSPs leading to DA have status `F', the attack fails 100\%; therefore, the value for these states is 0. Once the attacker reaches DA, there is no state transition as the attack ends immediately. For other states, we train the NN to learn the recursive relationship of Eq. (\ref{DP}). 
We select an initial state vector (actual state vector corresponding to a blocking plan; the state where some coordinates are `F', which are the NSPs blocked by the blocking plan and rest of the coordinates are `?') as shown in Eq. (\ref{attacker state vector}) and generate a batch of future states to train NN. Given an initial state, NN makes an optimal decision (according to the NN model) with $0.5$ probability or makes a random decision with $0.5$ probability to explore other possibilities. It is possible that the action performed by NN is not optimal; therefore, we employ randomness to explore other actions as well. After performing an action, we may have a distribution of future states and their transition probabilities. We select one of the future states weighted by its probability. Similarly, we keep moving to other states until we reach the base state. We train the NN to learn the recursive relationship between states and minimize the mean squared error (MSE) of estimated results of the value function.
Let $V(s;\theta)$ be the value predicted by the NN, where $s$ represents the input state vector, and $\theta$ be the model parameter. Ideally, $V(s;\theta)$ is exactly a DP function. The loss is computed as follow:
\begin{equation}
\label{loss}
Loss = \sum_{s \in S} \bigg(V(s;\theta) - \max_{\substack{a\in A(s)}} \,\Big(\sum_{\substack{s'}}{Pr(s'\,|\,s,a)\, V(s')}\Big)\bigg)^2
\end{equation}
where $S$ is the set of all states. It is impractical to compute the gradient for all states in one iteration; therefore, we adopt a common approach of performing gradient descent on a mini-batch of data to train the NN. As the NN value function gets better, there are higher chances that the generated states are optimal or near-optimal, which indicates that the attacker may go via these states to DA. In addition, given a deep enough neural network model and unlimited time, this will give us optimal results.

\subsection{Defender's Policy: Evolutionary Diversity Optimisation (EDO)}
The defender uses Evolutionary Diversity Optimisation to block $k$ block-worthy edges in order to minimise the strategic attacker probability of reaching DA. NN acts as a fitness function for EDO and the fitness function computes the attacker's probability of reaching DA, for a given blocking plan. The defender uses EDO to obtain a diverse set of defensive blocking plans to train NN, with an aim to improve the accuracy of trained NN for modelling the attacker. We have only considered the block-worthy edges for the defensive policy. The defensive state vector can be represented as: 
\begin{equation}
\label{def_state_vec}
\underbrace{< 0, 1, 0, 0, 1, 1, 0, 0>}_\text{Length of defensive state vector = Number of block-worthy edges} 
\end{equation}
where:
\begin{description}
\item `1' represents blocked edges
\item `0' represents non-blocked edges
\end{description}
\textbf{Evolutionary Diversity Optimisation.} We initially generate a random population $P$ of defensive blocking plans, and each blocking plan is a vector of size $|BW|$. The values in a state vector are either $1$ or $0$, and the number of $1$s is always equal to $k$ (defensive budget). We randomly select individuals from $P$ for mutation and crossover. We also draw a random number $x$ from Poisson distribution with a mean value equal to 1 and perform either mutation or crossover with a probability of $0.5$ to create new offspring. Our mutation and cross-over operators ensure that the number of blocked edges in an offspring does not exceed $k$; the mutation and crossover operations are discussed below:\\

\noindent \textit{\textbf{Mutation.}} We pick a random individual $p'$ from the population $P$, and flip $x$ 0s to 1s and $x$ 1s to 0s. For example, we pick a random individual $p'$ = <1,0,0,0,1,0,0,1> from the population $P$. To mutate $p'$, we flip two 0s to 1s and two 1s to 0s. New individual obtain is: <0,1,0,0,1,0,1,0>.

\noindent \textit{\textbf{Crossover.}} We pick two random individual state vector $p'$ and $p''$ from the population $P$ to crossover. We find $x$ coordinates where $p'$ has 0s on those coordinates, and $p''$ has 1s on those coordinates. For these coordinates, we change 0s in $p'$ to 1s and change 1s in $p''$ to 0s. We then find $x$ coordinates where $p'$ has 1s on those coordinates and $p''$ has 0s on those coordinates. We again change 1s to 0s and 0s to 1s. 

\noindent \textbf{\textit{Diversity Measure.}} After mutation and crossover, we add the new individual to $P$ if its fitness value is close to optimal (within an absolute difference of 0.1); otherwise, we reject the individual even if it is good for the diversity of population. We consider the diversity in terms of equal representation of block-worthy edges in the population. Let us say there are $\mu$ individuals in the population $P$ and each individual $p_j$ is represented as: 
\begin{equation*}
  p_j =  \big((bw_1;j), (bw_2;j), ..., (bw_{|BW|};j)\big), \;\;\;\;\;\;j \in \{1,...,\mu\}
\end{equation*}
For each block-worthy edge $bw_i$, $\,\,i \in \{1,...,|BW|\}$, let $c(bw_i)$ denotes the block-worthy edge count as the number of individuals out of $\mu$ who have blocked this edge. Therefore, we get block-worthy edge count vector $C(bw)$ as:
\begin{equation*}
   C(bw) = (c(bw_1), c(bw_2), ..., c(bw_{|BW|}))
\end{equation*}

We now compute the diversity vector $D$ of the population $P$ without individual $p_{j}$ as: 
\begin{equation*}
  D(C(bw)\backslash{p_j}) = C(bw) - p_j
\end{equation*}
\begin{equation*}
   D(C(bw)\backslash{p_j}) = \Big(c(bw_1)-(bw_1;j),..., c(bw_{|BW|})-(bw_{|BW|};j)\Big)
\end{equation*}
where $D(C(bw)\backslash{p_j})$ represents the diversity of population without individual $p_{j}$. In order to maximize the blocked edge diversity, we aim to minimize the $SortedD(C(bw)\backslash{p_j})$ in the lexicographic order, where sorting is done in descending order.

\begin{flalign*}
\text{\textit{Sorted}} D(C(bw)\backslash{p_j})=  sort \Big(\big(c(bw_1)-(bw_1;j)\big),  .......,\big(c(bw_{|BW|})-(bw_{|BW|};j)\big)\Big)
\end{flalign*}

In this way, we determine the diversity \textit{SortedD}, of the population without each individual in the population. We then select the individual $r$, removal of which leads to maximum diversity (minimum $SortedD(C(bw)\backslash{p_r})$) in population. We reject $r$ if it contributes least to diversity and its fitness value is not close to optimal. One exception is that if the newly introduced individual is the best-performing individual, then we remove the individual with the worst fitness value instead. In this way, we get a diverse set of defensive blocking plans via EDO.\\

\noindent \textit{\textbf{Converting defensive blocking plan to actual state vector.}} For each defensive blocking plan vector (Eq. (\ref{def_state_vec})), we first need to convert it to the \textit{actual state vector} (Eq. (\ref{attacker state vector})) in order to determine the attacker's chances of reaching DA (corresponding to this blocking plan). "Actual state vector" is the state vector used in the attacker's MDP. For each block-worthy blocked edge $bw$ in the defensive state vector, we first determine the number of NSPs blocked by this $bw$ edge and then change the status of those NSPs to `F', keeping the status of rest to `?'.\\

\noindent \textbf{Overall approach: NNDP-EDO.} Initially, we have a NN which is highly inaccurate (attacker's policy). We generate a diverse set of blocking plans as training data for NN using EDO (Defender's policy), and NN acts as an efficient fitness function for EDO. Notably, the exact fitness function is $\#P$-hard to compute, therefore, we use NN as a fitness function. We convert the blocking plans to the actual state vectors and in each training epoch select a random blocking plan out of the population to train NN, which we call an initial state vector. We use the state transition process to determine future states set, transition probabilities and admissible action set. We then train NN using Eq. (\ref{loss}) on the future state set to approximate the value function, i.e., NN outputs attacker's probability of reaching DA with the given a blocking plan. We repeatedly train NN on a diverse blocking plans aiming to improve the accuracy of NN for modelling the attacker. We repeatedly go back and forth between training NN and EDO processes for many rounds to get a well trained NN that can act as an efficient fitness function for EDO. In the last round, we train NN once again on the population obtained from the last round of EDO. EDO prevents the NN model from getting stuck into local optima too early, especially in the early phases of the training when the value function is highly inaccurate. In addition, not all states are important for the attacker; therefore, we let the value of states that are referenced by the attacker's optimal decision path, be more accurate by training NN on these states.

\section{Experimental Results}
We discuss the effectiveness of the proposed approach by conducting exhaustive experiments on various AD attack graphs. 
We conducted all the experiments on a high-performance cluster server with Intel Gold 6148/6248 CPUs. All trials are executed using 1 CPU and 1 Core. Notably, we conducted the experiments on a very large scale, and it took us 141.47 days of computing hours to run all the experiments. We allocated 180 CPUs from the high performance cluster to run our 180 trials in parallel, therefore, our experiment completed in 1 day. We implemented the code in PyTorch. 

\begin{table}[h!] 
\caption{Summary of synthetic original AD attack graphs.}
\label{dataset}
\renewcommand{\arraystretch}{1}
\centering 
\begin{tabular}{ccc} \hlineB{2} 
\textbf{{AD attack graph} }& \textbf{{Nodes}} & \textbf{{Edges}}\\ \hlineB{2}
R500 & 1493 & 3456 \\
R1000 & 2996 & 8814\\
R2000 & 5997 & 18795\\\hlineB{2}
\end{tabular} 
\end{table}

\subsection{Synthetic AD attack graphs} 
We generated synthetic R500, R1000, R2000 AD attack graphs using DBCREATOR, where 500, 1000 and 2000 are the number of computers. In addition, we only consider three types of edges by default present in BLOODHOUND; HasSession, AdminTo, and MemberOf. Table \ref{dataset} presents the summary of original attack graphs, where nodes in original graphs include computers, user accounts and security groups. We pick 40 nodes that are farthest away from DA (in terms of number of hops) and randomly select 20 nodes as entry nodes. An edge$(i,j)$ is set to be blockable with probability $L$:
\begin{equation*}
  L =  \frac{\text{Min number of hops between edge$(i,j)$ and DA}}{\text{Max number of hops between edge$(i,j)$ and DA}}
\end{equation*}
In this way, the farthest edges from DA are more likely to be blockable, which are less important. 
We pre-process the original attack graphs to obtain the condensed graphs, which only contains splitting nodes. For instance, in R500 original graph, there are 7 DA, 1493 nodes and 3456 edges. We merge the 7 DA into one destination node and remove all the outgoing edges from DA (once the attacker reaches DA, the attack ends). As the attacker will never use the incoming edges to entry nodes, we remove them. We also remove the nodes with no incoming edges. Out of 1493 nodes, only 105 can reach DA; we remove all the nodes that can not reach DA. In addition, splitting nodes are connected via non-splitting paths, therefore, we consider a non-splitting path as a single edge. In this way, we obtain a condensed graph, which is much smaller than the original AD graph.

\subsection{Training Details}
In our approach, we use a simple fully connected Neural Network, and a ReLU activation function follows each NN layer. For the R500 graph, a small NN with 4 fully connected layers is used, whereas, for R1000 and R2000, we use a NN with 10 layers. The size of each layer is 256 and the last layer of NN is followed by a sigmoid activation function that maps the model output to a value between 0 and 1 (the attacker's chances of reaching DA). We train the NN in a batch of 16 states. We use the mean squared error to compute the loss, and train the parameters using Adam Optimizer with a learning rate of 0.001. The model is trained for 500 epochs in each round. 
We set the defender budget to 5. We generate a population of 100 defensive blocking plans in 10000 iterations and perform mutation or crossover with a probability of 0.5. We repeat the combined process (training NN and generating blocking plans using EDO) for 100 rounds. We perform experiments with a seed from 0 to 9 for 10 separate trials (we conduct experiments on 10 different AD graphs with different entry nodes and different blockable edges).

\subsection{Baselines}
We compare our proposed approach with a combination of various attacking and defensive policies, and the details are described below:
\begin{enumerate}[leftmargin=*]
    \item \textit{\textbf{NNDP-EDO (Proposed).}} We proposed NNDP approach to approximate the attacking policy and EDO for defensive policy. The defensive plan that contributes least to the diversity is rejected, and the fitness value is evaluated using NNDP.
    \item \textit{\textbf{NNDP-EDO+DP.}} NNDP is used to approximate the attacking policy and EDO for defensive policy. However, instead of using NNDP to evaluate the success rate of best blocking plan (as in our proposed approach), we use accurate DP to determine the value of the defensive plan.
    \item \textit{\textbf{Exact solution.}} Dynamic Program is used as attacker's policy, and defensive blocking plan is obtained by exhaustively trying each in order to get the best plan.
    \item \textit{\textbf{NNDP-VEC.}} In this approach, NNDP is used to approximate the attacking policy, and Value-based Evolutionary Computation (VEC) is used to design the defensive policy. In VEC, the blocking plans with the worst fitness values are rejected, and the fitness value is evaluated using NNDP.
    \item \textit{\textbf{NNDP-Greedy.}} The attacker uses NNDP to approximate the attacking policy and defender adopts a greedy approach to design the defensive policy. The defender greedily blocks single best block-worthy edge to minimise the attacker's chances of reaching DA using NNDP. The defender repeats this for $k$ times.
\end{enumerate}
Notably, NNDP-EDO+DP and Exact solution use DP to evaluate the blocking plan; however, it is infeasible for DP to process large graphs. Therefore, for R500 graph, we compare our proposed approach NNDP-EDO, with NNDP-EDO+DP and Exact solution. For larger graphs R1000 and R2000, we compare our approach with the NNDP-VEC and NNDP-Greedy to determine our defensive strategy's effectiveness. The trained NNDP may not give us the accurate values of success rate for the defensive plan; therefore, we use Monte Carlo simulations to determine the effectiveness of the defensive plan. Moreover, in order to investigate the impact of correlation between failure rate $p_f$ and detection rate $p_d$ of edges on the success rate of attacker, we have considered three types of distribution; independent, positive correlation and negative correlation. In independent distribution, $p_f$ and $p_d$ are mutually independent, and we set $p_f$ and $p_d$ based on independent uniform distribution from 0 to 0.2. Positive correlation between $p_f$ and $p_d$ indicates that both have a steady relationship in the same direction. Negative correlation between $p_f$ and $p_d$ indicates an inverse relationship; one decreases as the other increases. We use multivariate normal distribution to get $p_f$ and $p_d$ for positive and negative correlation between an edge $e$ as:
\begin{equation*}
p_{d(e)}, p_{f(e)} = \text{multivariate normal (mean, cov)}
\end{equation*}
For positive correlation, mean = $[0.1, 0.1]$ and cov = $[[0.05^2, 0.5\times0.05^2], [0.5 \times 0.05^2, 0.05^2]]$. For negative correlation, mean = $[0.1, 0.1]$ and cov = $[[0.05^2, -0.5\times0.05^2], [-0.5 \times 0.05^2, 0.05^2]]$.

\begin{table*}[t!] 
\caption{Comparison of Success rate(\%) on R500 AD attack graph.}
\label{r500}
\renewcommand{\arraystretch}{1}
\centering 
\begin{tabular}{llllll} \hlineB{2} 


\textbf{AD Graph} & \textbf{Approach} &\textbf{Independent} & \textbf{Positive} & \textbf{Negative} & \textbf{Average}\\ \hlineB{2} 

      & NNDP-EDO (Proposed) & 87.96\% & 84.77\% & 84.5\% & 85.74\%\\
R500 & NNDP-EDO + DP & 90.49\% & 84.8\% & 84.46\%  & 86.58\%  \\
      & Exact solution & 89.73\%  & 84.78\% & 84.45\% & 86.32\% \\\hlineB{2} 
\end{tabular} 
\end{table*}

\begin{table*}[t!] 
\caption{Comparison of Success rate(\%) on R1000 AD attack graph.}
\label{r1000}
\renewcommand{\arraystretch}{1}
\centering 
\resizebox{\textwidth}{!}{\begin{tabular}{lllllllll} \hlineB{2} 

\multicolumn{1}{c}{\textbf{{}}}& \multicolumn{1}{c}{\textbf{{}}}&  \multicolumn{4}{c}{\textbf{{Success Rate}}} & \multicolumn{3}{c}{\textbf{{Time(s)}}} \\  \cmidrule(lr){3-6}  \cmidrule(lr){7-9}

\textbf{AD Graph} & \textbf{Approach} &\textbf{Independent} & \textbf{Positive} & \textbf{Negative} &  \textbf{Average} & \textbf{Independent} & \textbf{Positive} & \textbf{Negative}\\ \hlineB{2} 

      & NNDP-EDO (Proposed) & 42.69\% & 42.44\% & 41.29\% & 42.14\% & 39410.85 & 60263.28  & 64093.08 \\
R1000 & NNDP-VEC & 43.19\% & 44.7\% & 41.24\% & 43.04\% & 39815.72 & 61572.04  & 57708.75 \\
      & NNDP-Greedy & 53.89\%  & 52.86\% & 50.31\% & 52.35\% & 38413.97 & 64555.83 & 67436.33 \\\hlineB{2} 
\end{tabular}}
\end{table*}

\begin{table*}[t!] 
\caption{Comparison of Success rate(\%) on R2000 AD attack graph.}
\label{r2000}
\renewcommand{\arraystretch}{1}
\centering 
\resizebox{\textwidth}{!}{\begin{tabular}{lllllllll} \hlineB{2} 

\multicolumn{1}{c}{\textbf{{}}}& \multicolumn{1}{c}{\textbf{{}}}&  \multicolumn{4}{c}{\textbf{{Success Rate}}} & \multicolumn{3}{c}{\textbf{{Time(s)}}} \\  \cmidrule(lr){3-6}  \cmidrule(lr){7-9}

\textbf{AD Graph} & \textbf{Approach} &\textbf{Independent} & \textbf{Positive} & \textbf{Negative} &  \textbf{Average} & \textbf{Independent} & \textbf{Positive} & \textbf{Negative}\\ \hlineB{2} 

      & NNDP-EDO (Proposed) & 31.07\% & 33.9\% & 35.56\% & 33.51\% & 27291.01 & 65488.9 & 59843.9 \\
R2000 & NNDP-VEC & 33.23\% & 36.45\% & 34.19\% & 34.62\% & 26838.6 & 67079.8 & 57160.1 \\
      & NNDP-Greedy & 38.32\%  & 42.69\% & 39.02\% & 40.01\% & 24478.4 & 65620.3 & 57412.8 \\\hlineB{2} 
\end{tabular}}
\end{table*}

\subsection{Results}
In our results, “Success Rate” represents the attacker's chances of successfully reaching DA before getting detected (given attacker's policy) under the defensive blocking plan from the defender's policy. “Time (s)” is the number of seconds per trial. 
We simulate the attacker's policy (NNDP) on the best defensive blocking plan (predicted by NNDP) using Monte Carlo simulations over 100000 runs to determine attacker's chances of reaching DA. We performed all the experiments ten times with a seed from 0 to 9, and the average results over all seeds are presented as final results.

\noindent \textit{\textbf{Results on R500.}} Table \ref{r500} presents the success rate of the proposed approach NNDP-EDO and other baselines on synthetic R500 graph under various distributions. The attacker's average success rate is  87.96\% when simulated using Monte Carlo over 100000 runs with our proposed defense EDO and NNDP (under independent distribution). Moreover, the exact actual success rate for our proposed defense when evaluated using NNDP-EDO+DP is 90.49\%. This indicates that for a blocking plan from EDO, our trained NNDP generates an error of 2.53\% in the success rate. For exact optimal defense, the attacker's success rate is 89.73\%, which indicates that our proposed defense is 0.76\% (less than 1\%) away from the optimal. Similarly, the proposed defense NNDP-EDO performs near-optimal under positive correlation; NNDP-EDO believes the best defense has a success rate of 84.77\%, but in reality, the accurate success rate of the proposed defense is 84.8\%, which is slightly worse than the exact solution 84.78\%. In negative correlation as well, NNDP-EDO defense is nearly as effective as optimal.

\noindent \textit{\textbf{Results on R1000.}} Table \ref{r1000} presents the results for R1000 AD attack graph. It is impossible to determine the exact attacker's success rate for R1000 graph using DP; therefore, we use NNDP to approximate the attacker's policy. Table \ref{r1000} shows that, on an average, the proposed EDO based defense NNDP-EDO, leads to the best defensive policy as compared to the other defense (NNDP-VEC and NNDP-Greedy). For instance, under independent distribution, the attacker's success rate is 42.69\% with EDO defensive policy; however, the success rate increases to 43.19\% and 53.89\% when facing VEC defense and Greedy defense, respectively. Under positive correlation, NNDP-EDO is the best defense among the three and leads to minimum attacker's success rate. However, under negative correlation, the VEC defense is slightly better than EDO. The results show that overall EDO's best defense has an average success rate of 42.14\%; VEC best defense has a success rate of 43.04\%, which is slightly worse than EDO. The greedy defense has an average success rate of 52.35\%, which is far worse than VEC.

\noindent \textit{\textbf{Results on R2000.}} The results for R2000 AD attack graphs in Table \ref{r2000} show that EDO based defence outperforms VEC and Greedy based defense in terms of average success rate. With the EDO defense (under independent distribution), there are only 31.07\% chances of attacker's reaching DA; however, with the Greedy defense, the attacker's success rate increases to 38.32\%. EDO performs better than VEC and Greedy in both independent and positive correlation; however, VEC performs slightly better than EDO in negative correlation. On an average best defense from EDO has an average success rate of 33.51\%; VEC has success rate of 34.62\%, slightly worse than EDO and Greedy has an average success rate of 40.01\%, which is very high as compared to EDO defense.

\noindent \textit{\textbf{Discussion.}}The results show that the proposed defense NNDP-EDO is highly effective. We have exact optimal results for R500 attack graph, and on average, our proposed approach is less than 1\% away from the optimal defense. In addition, the proposed attacker's policy NNDP approximates the success rate for the defense with high accuracy and incurs a small error of 2.53\%. This shows that EDO trains NNDP very effectively, and trained NNDP acts as a very efficient fitness function for EDO. It is impossible for large R1000 and R2000 graphs to run the exact DP evaluation function; therefore, we simulate the defense using Monte Carlo simulation to get attacker's success rate. The results show that the proposed approach NNDP-EDO is better than others, and overall, the best defense from EDO has an average success rate of 42.14\% for R1000 and 33.51\% for R2000, which is far less than other defensive approaches. In addition, the results show that the Value-based Evolutionary Computation proves to be a better defense than the greedy defense. For R1000 and R2000 graphs, 6/180 trails (around 3\% ) ended up with unconverged NN training, i.e., after 500$\times$100 epochs, the cost function remains to be large.

\section{Conclusion}
This paper investigated a Stackelberg game model on an Active Directory attack graph between an attacker and a defender. The defender aims to block a number of edges to minimise the attacker's probability of reaching DA; however, the attacker aims to maximise their chances of reaching DA. We first proved that both the attacker's and defender's problems are $\#P$-hard. We proposed Evolutionary Diversity Optimisation to solve the defender's problem, and the Neural Network acts as an efficient fitness function for defender's blocking plans. The experimental results showed that the proposed Evolutionary Diversity Optimisation based defensive policy is highly effective and our approach can solve large AD attack graph problems, which are intractable for conventional Dynamic Program. For R500 AD attack graph, our proposed approach is less than 1\% away from the optimal defense.

\section*{Acknowledgements}
This work has been supported by South Australian Government through the Research Consortium ``Unlocking Complex Resources through Lean Processing”,  Australian Research Council through grants DP190103894 and FT200100536, partially supported by the ``Cyber NGT – Provable Network Security” grant and with supercomputing resources provided by the Phoenix HPC service at the University of Adelaide.

\bibliographystyle{unsrtnat}
\bibliography{references}

\end{document}